\documentclass[10pt,twocolumn,letterpaper]{article}	

\usepackage{cvpr}
\usepackage{times}
\usepackage{epsfig}
\usepackage{graphicx}
\usepackage{amsmath}
\usepackage{amssymb}
\usepackage{graphicx}
\usepackage{subcaption}
\usepackage{setspace}

\usepackage{calrsfs}
\DeclareMathAlphabet{\pazocal}{OMS}{zplm}{m}{n}
\newcommand{\Lb}{\pazocal{L}}

\usepackage[pagebackref=true,breaklinks=true,letterpaper=true,colorlinks,bookmarks=false]{hyperref}

\cvprfinalcopy 

\ifcvprfinal\fi

\begin{document}
\pagenumbering{gobble}
\title{Classification Driven Dynamic Image Enhancement}
\author{
    {  Vivek Sharma$^{1}$, Ali Diba$^{2}$, Davy Neven$^{2}$, Michael S. Brown$^{3}$,  Luc Van Gool$^{2,4}$, and Rainer Stiefelhagen$^{1}$}\\
    {\normalsize {$^{1}$CV:HCI, KIT, Karlsruhe, $^{2}$ESAT-PSI, KU Leuven, $^{3}$York University, Toronto, and $^{4}$CVL, ETH Z\"{u}rich}} \\ 
     \tt\small \{firstname.lastname\}@kit.edu, \{firstname.lastname\}@esat.kuleuven.be, mbrown@eecs.yorku.ca
 }


\maketitle
\begin{abstract}
Convolutional neural networks rely on image texture and structure to serve as discriminative features to classify the image content.  Image enhancement techniques can be used as preprocessing steps to help improve the overall image quality and in turn improve the overall effectiveness of a CNN. Existing image enhancement methods, however, are designed to improve the perceptual quality of an image for a human observer.  In this paper, we are interested in learning CNNs that can emulate image enhancement and restoration, but with the overall goal to improve image classification and not necessarily human perception.  To this end, we present a unified CNN architecture that uses a range of enhancement filters that can enhance  image-specific details via end-to-end dynamic filter learning. We demonstrate the effectiveness of this strategy on four challenging benchmark  datasets for fine-grained, object, scene, and texture classification: CUB-200-2011, PASCAL-VOC2007,  MIT-Indoor, and DTD. Experiments using our proposed enhancement show promising results on all the datasets. In addition, our approach is capable of improving the performance of all generic CNN architectures.
\end{abstract}

\section{Introduction}

Image enhancement methods are commonly used as preprocessing steps that are applied to improve the visual quality of an image before higher level-vision tasks, such as classification and object recognition~\cite{sharma_cic2,sharma_cic1}.  Examples include  enhancement to remove the effects of blur, noise, poor contrast, and compression artifacts -- or to boost image details. Examples of such enhancement methods include Gaussian smoothing, anisotropic diffusion, weighted least squares (WLS), and bilateral filtering. Such enhancement methods  are not simple filter operations (e.g., 3$\times$3 Sobel filter), but often involve complex optimization. In practice, the run time for these methods is expensive and can take seconds or even minutes for high-resolution images. 

Several recent works have shown that convolutional neural networks (CNN)~\cite{ayan_2016_2,koltun,liu,deepclass,deepaware,yan} can successfully emulate a wide range of image enhancement by training on input and target output image pairs. These CNNs often have a significant advantage in terms of run-time performance.   The current strategy, however, is to train these CNN-based image filters to approximate the output of their non-CNN counterparts.  

In this paper, we propose to extend the training of CNN-based image enhancement to incorporate the high-level goal of image classification.  Our contribution is a method that jointly optimizes a CNN for enhancement and image classification. We achieve this by adaptively enhancing the features on an image basis via dynamic convolutions, which enables the enhancement CNN to selectively enhance only those features that lead to improved image classification.

\begin{figure}[t]
\centering
{\includegraphics[width=1\columnwidth]{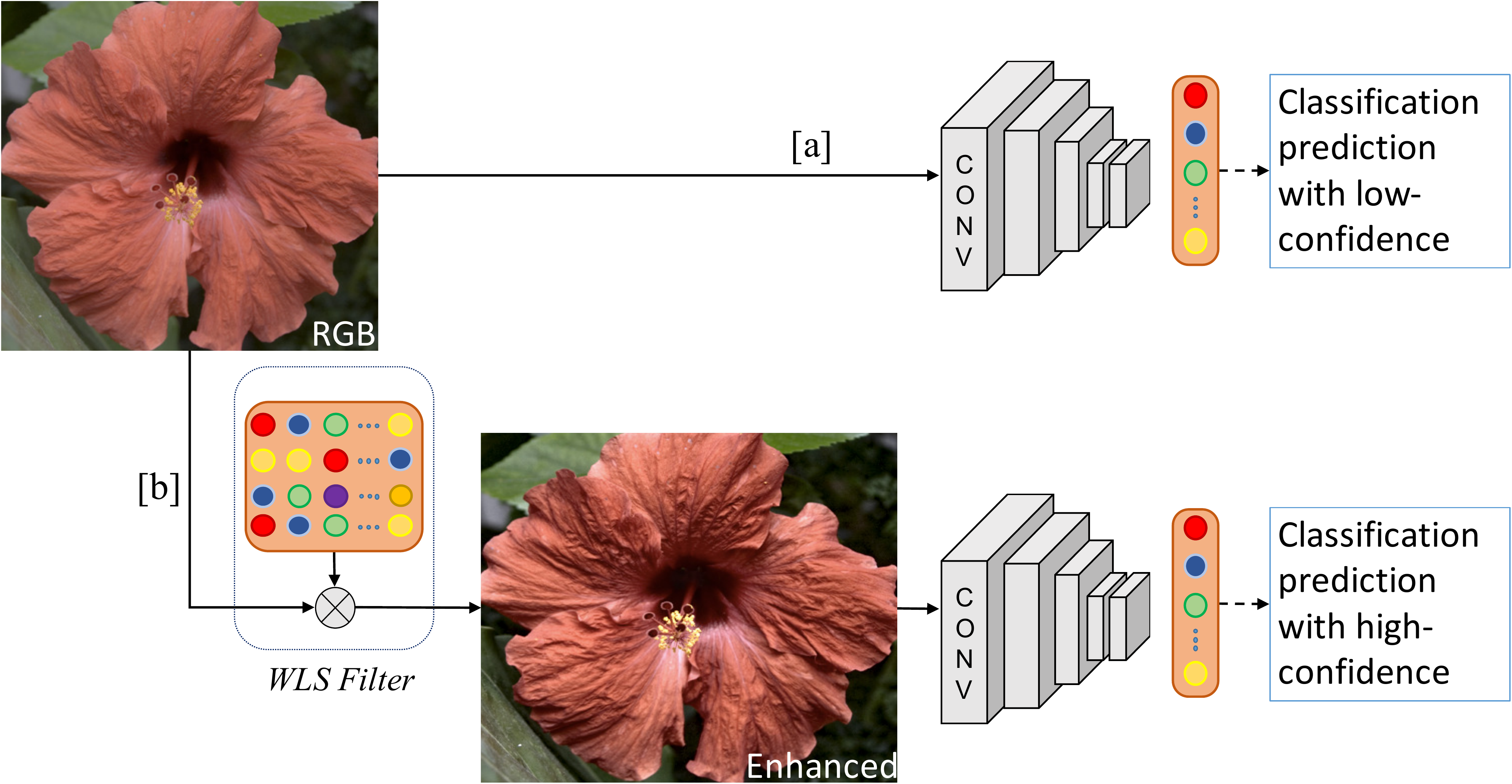} } 
\caption{\small{Overview of the proposed unified CNN architecture using enhancement filters to improve classification tasks. Given an input RGB image, instead of directly applying the CNN on this image ([a]), we first enhance the image details by convolving the input image with a WLS filter (see Sec.~\ref{subsec:app1}), resulting in improved classification with high confidence ([b]).}}\vspace{-0.5cm} 
\label{fig:front}
\end{figure} 

Since we understand the critical role of selective feature enhancement, we propose to use the dynamic convolutional layer (or dynamic filter)~\cite{dfn} to dynamically enhance the \textit{image-specific} features with a classification objective (see Fig.~\ref{fig:front}). Our work is inspired by ~\cite{dfn}. However, while~\cite{dfn} applies the dynamic convolutional module to transform an angle into a filter (steerable filter) using input-output image pairs, we used the same terminology as in~\cite{dfn}. The dynamic filters are a function of the input and therefore vary from one sample to another during train/test time, which means when the image enhancement is done in an \textit{image-specific} way to enhance the texture patterns or sharpen edges for discrimination. Specifically, our network learns the amount of various enhancement filters that should be applied to an input image, such that the enhanced representation provides better performance in terms of classification accuracy. Our proposed approach is evaluated on four challenging benchmark datasets for fine-grained, object, scene, and texture classification respectively: CUB-200-2011~\cite{cubdataset}, PASCAL-VOC2007~\cite{pascalvocdataset},  MIT-Indoor~\cite{mitdataset}, and DTD~\cite{dtddataset}. We experimentally show that when CNNs are combined with the proposed dynamic enhancement technique (Sec.~\ref{subsec:app1} and \ref{subsec:app3}), one can consistently improve the classification performance  of vanilla CNN architectures on all the datasets. In addition, our experiments demonstrate the full capability of the proposed method, and show promising results in comparison to the state-of-the-art.

The remainder of this paper is organized as follows. Section~\ref{sec:related} overviews related work.  Section~\ref{sec:app} describes our proposed enhancement architecture. Experimental results and their analysis are presented in Sections~\ref{sec:exp}. Finally, the paper is concluded in Section~\ref{sec:conc}.

\begin{figure*}[htb]
 \centering
 \includegraphics[width=2\columnwidth]{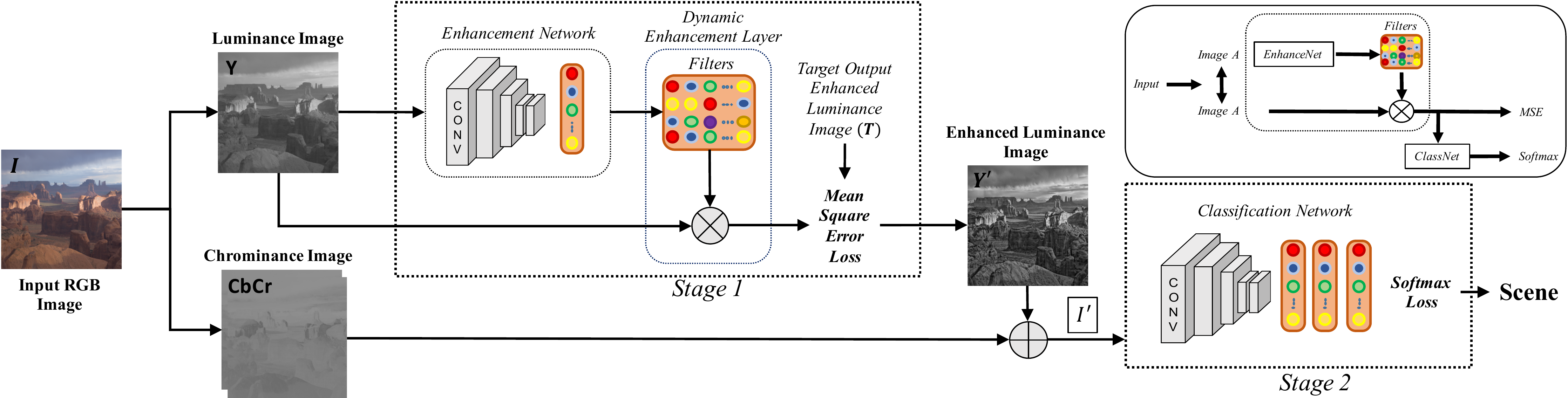}\vspace{-0.25cm}
 \caption{\small{\textbf{Dynamic enhancement filters}. Input to the network are input-output image pairs, as well as image class labels for training. In this architecture, we \textbf{learn a single enhancement filter} for each enhancement method individually. The model operates on the luminance component of RGB color space. The enhancement network (i.e., filter-generating network) generates dynamic filter parameters that are \textit{sample-specific} and conditioned on the input of the enhancement network, with the overall goal to improve image classification. The figure in the upper-right corner shows the whole pipeline workflow.}}\vspace{-0.5cm}
  \label{fig:2}
 \end{figure*}

\section{Background and Related Work} \label{sec:related}

\noindent
Considerable progress has been seen in the development for removing the effects of blur~\cite{ayan_2016_2}, noise~\cite{deepclass}, and compression artifacts~\cite{xu2014deep} using CNN architectures. Reversing the effect of these degradations in order to obtain sharp images is currently an active area of research~\cite{ayan_2016_2,li2016deep,deepaware}. The investigated CNN frameworks~\cite{ayan_2016_2,koltun,derain,li2016deep,liu,deepclass,deepaware,yan} are typically built on simple strategies to train the networks by minimizing a global objective function using input-output image pairs. These frameworks encourage the output to have a similar structure with the target image. After training the CNN, a similar approach to transfer details to new images has been proposed~\cite{deepaware}. These frameworks act as a filter that are specialized for a specific enhancement method.

For example, Xu et al.~\cite{deepaware} learn a CNN architecture to approximate existing edge-aware filters from input-output image pairs. Chen et al.~\cite{koltun} learn a CNN that approximates end-to-end several image processing operations using a parameterization that is deeper and more context-aware. Yan et al.~\cite{yan} learn a CNN to approximate image transformations for image adjustment. Fu et al.~\cite{derain} learn a CNN architecture to remove rain streaks from an image. For CNN training, the authors use rainy and clean image detail layer pairs rather than the regular RGB images.  Li et al.~\cite{li2016deep} propose a learning-based joint filter using three CNN architectures. In Li et al.'s work, two sub-networks take target and guidance images, while the third-network selectively transfers the main content structure and reconstructs the desired output.  Remez et al.~\cite{deepclass} propose a fully convolutional CNN architecture to do image denoising using image prior-that is, class-aware information. The closest work to ours is by Chakrabarty et al.~\cite{ayan_2016_2} and Liu et al.~\cite{liu}. Chakrabarty et al. propose a CNN architecture to predict the complex Fourier coefficients of a deconvolution filter which is applied to individual image patches for restoration. Liu et al. use CNN+RNNs to learn enhancement filters; here we use CNNs only for learning filters. Our methods produce one representative filter per method, while they produce four-way directional propagation filters per method. Like others, their work is meant for low-level vision tasks similar to~\cite{ayan_2016_2,koltun}, while our goal is enhancement for classification. In contrast to these prior works, our work differs substantially in scope and technical approach. Our goal is to approximate different image enhancement filters with a classification objective in order to selectively extract informative features from enhancement techniques to improve classification, not necessarily approximating the enhancement methods.

Similar to our goal are the works~\cite{empirical,dodge,karahan,peng2016fine,ullman,ayan_2016}, where the authors also seek to ameliorate the degradation effects for accurate classification.  Dodge and Karam~\cite{dodge} analyzed how blur, noise, contrast, and compression hamper the performance of ConvNet architectures for image classification. Their findings showed that: (1) ConvNets are very sensitive to blur because blur removes textures in the images; (2) noise affects the performance negatively, though deeper architectures' performance falls off slower; and (3) deep networks are resilient to compression distortions and contrast changes.  A study by Karahan et al.~\cite{karahan} reports similar results for a face-recognition task.  Ullman et al.~\cite{ullman} showed that minor changes to the image, which are barely perceptible to humans, can have drastic effects on computational recognition accuracy. Szegedy et al.~\cite{Zaremba} showed that applying an imperceptible non-random perturbation can cause ConvNets to produce erroneous prediction. 

To help to mitigate these problems,  Costa et al.~\cite{empirical} designed separate models  specialized for each noisy version of an augmented training set. This  improved the classification results for noisy data to some extent. Peng et al.~\cite{peng2016fine} explored the potential of jointly training on low-resolution and high-resolution images in order to boost performance on low-resolution inputs. Similar to~\cite{peng2016fine} is Vasijevic et al.'s~\cite{ayan_2016} work, where the authors augment the training set with degradations and fine-tune the network with a diverse mix of different types of degraded and high-quality images to regain much of the lost accuracy on degraded images. In fact, with this approach the authors were able to learn to generate a degradation (particularly blur) invariant representation in their hidden layers. 

In contrast to previous works, we use high-quality images that are free of artifacts, and jointly learn ConvNet to enhance the image for the purpose of improving classification.

\section{Proposed Method} \label{sec:app}

As previously mentioned, our aim is to learn a dynamic image enhancement network with the overall goal to improve classification, and not necessarily approximating the enhancement methods specifically.  To this end, we propose three CNN architectures described in this section.   

Our first architecture is proposed to learn a single enhancement filter for each enhancement method in an end-to-end fashion (Sec.~\ref{subsec:app1}) and by end-to-end we mean each image will be enhanced and recognized in one unique deep network with dynamic filters.  Our second architecture uses pre-learned enhancement filters from the first architecture and combines them in a weighted way in the CNN. There is no adaptation of weights of the filters (Sec.~\ref{subsec:app2}). In our third architecture, we show end-to-end joint learning of multiple enhancement filters (Sec.~\ref{subsec:app3}). We also combine them in a weighted way in the CNN. All these setups are jointly optimized with a classification objective to selectively enhance the image feature quality for improved classification.  In the network training, image-level class labels are used, while for testing the input image can have multiple labels.

\subsection{Dynamic Enhancement Filters}\label{subsec:app1}
In this section we describe our model to learn representative enhancement filters for different enhancement methods from input and target output enhanced image pairs in the end-to-end learning approach with a goal to improve classification performance. Given an input RGB image  $I$, we first transform it into the luminance-chrominance $YCbCr$ color space. Our enhancement method operates on the luminance component~\cite{luminance} of the RGB image.  This allows our filter to modify the overall tonal properties and sharpness of the image without affecting the color. The luminance image $Y \in \mathbb{R}^{h \times w}$ is then convolved with an image enhancement method $E: Y \rightarrow T$, resulting in an enhanced target output luminance image $T \in \mathbb{R}^{h \times w}$,  where $h$, and $w$ denote the height and width in the input $Y$ respectively. We generate target images for a range of enhancement methods $E$ as a preprocessing step (see Section.~\ref{subsec:gt} for more details). For filter generation, we explicitly use a dataset of only one enhancement method at a time for learning the transformation. The scheme is illustrated in Figure~\ref{fig:2}.

\textbf{First stage (enhancement stage):} The enhancement network (EnhanceNet) is inspired by~\cite{dfn,stn,dcl}, and is composed of convolutional and fully-connected layers. The EnhanceNet maps the input to the filter. The enhancement network  takes the  one channel luminance image $Y$ and outputs filters $f_{\Theta}$, $\Theta \in \mathbb{R}^{s \times s \times n}$, where $\Theta$ is the parameters of the transformation generated dynamically by the enhancement network, $s$ is the filter size, and $n$ is the number of filters, being equal to $1$ for a single generated filter meant for one channel luminance image. The generated filter is applied to the input image $Y(i,j)$ at every spatial position ($i$,$j$) to output predicted image $Y^{'}(i,j)=f_{\Theta}(Y(i,j))$ with $Y^{'} \in \mathbb{R}^{h \times w}$. The filters are \textit{image-specific}, and are conditioned on $Y$. For generating the enhancement filter parameters $\Theta$, the network is trained using mean squared error ($MSE$) between the target image $T$ and the network's predicted output image $Y^{'}$. Note that, the parameters of the filter are obtained as the output of a EnhanceNet that maps the input to a filter and therefore vary from one sample to another.  To compare the reconstruction image $Y^{'}$ with the ideal $T$, we use MSE loss as a measure of image quality, although we note that more complex loss functions could be used~\cite{dosovitskiy2016generating}.

The chrominance component is then recombined, and the image is transformed back into RGB, $I^{'}$.  We found that the filters learned the expected transformation and applied the correct enhancement to the image. Figure~\ref{fig:5} shows qualitative results with dynamically enhanced image textures.

\textbf{Second stage (classification stage):}
The predicted output image $I^{'}$ from Stage 1 is fed as an input to the classification network (ClassNet). As the classification network (e.g., Alexnet~\cite{alexnet}) has fully-connected layers between the last convolutional layer and the classification layer, the parameters of the fully-connected layer and $C$-way classification layer are learned when fine-tuning a pre-trained network. 

\textbf{End-to-end learning:} The Stage 1-2 cascade with two loss functions - $MSE$ ($enhancement$) and softmax-loss $\Lb$ ($classification$) - enables joint optimization by end-to-end  propagation of gradients  in both ClassNet and EnhanceNet using SGD optimizer. The total loss function of the whole pipeline is given by:
\begin{equation}
\begin{split}
Loss_{Filters} = MSE(T,Y^{'})+\Lb(\mathbf{P},\mathbf{y}) \quad \quad \quad \\ 
P_{q} = \frac{exp(a_{q})}{\sum_{r=1}^{C}exp(a_{r})}, \Lb(\mathbf{P},\mathbf{y}) = -\sum_{q=1}^{C}y_{q}log(P_{q})   \quad
\end{split}
\label{eq:2}
\end{equation}
where $a$ is the output of the last fully-connected layer of ClassNet that is fed to a $C$-way softmax function,  $\mathbf{y}$ is the vector of true labels for image $I$, and $C$ is the number of classes.

We fine-tune the whole pipeline until convergence, thus leading to learned enhancement filters in the dynamic enhancement layer.  The joint optimization allows the loss gradients from the ClassNet to also back-propagate through the EnhanceNet, making the filter parameters also optimized for classification.

\begin{figure*}[htb]
\centering
\includegraphics[width=2\columnwidth]{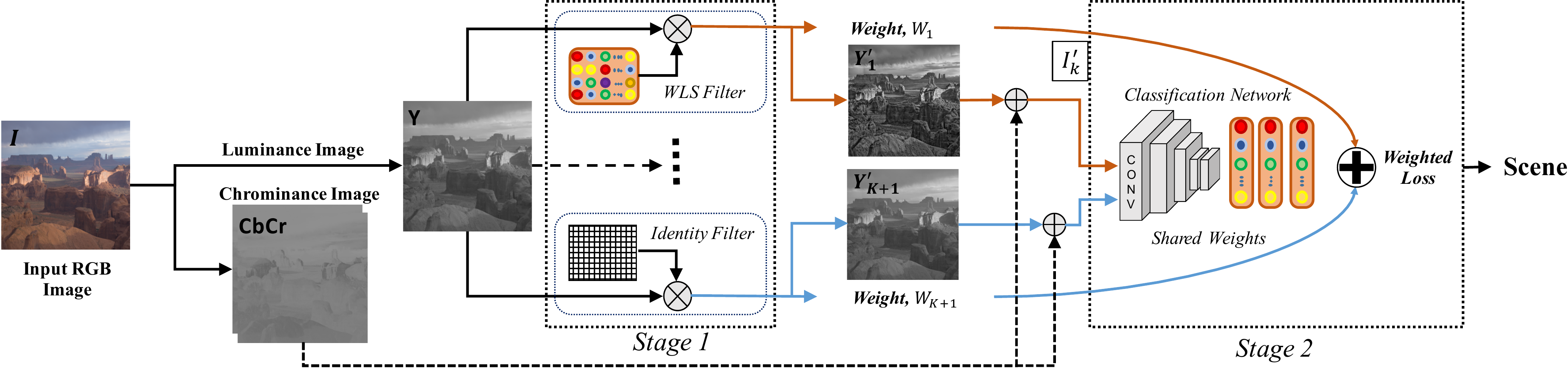}
\caption{ \small{\textbf{(Stat-CNN)} In this architecture, we use pre-learned filters from Sec.~\ref{subsec:app1} (Figure~\ref{fig:2}) for image enhancement  and the original image. The individual softmax scores are combined in a weighted way in the CNN. There is \textbf{no adaptation} of weights of the filters.}} \vspace{-0.2cm}
\label{fig:3}
\end{figure*} 
 
\begin{figure*}[htb]
\centering
\includegraphics[width=2\columnwidth]{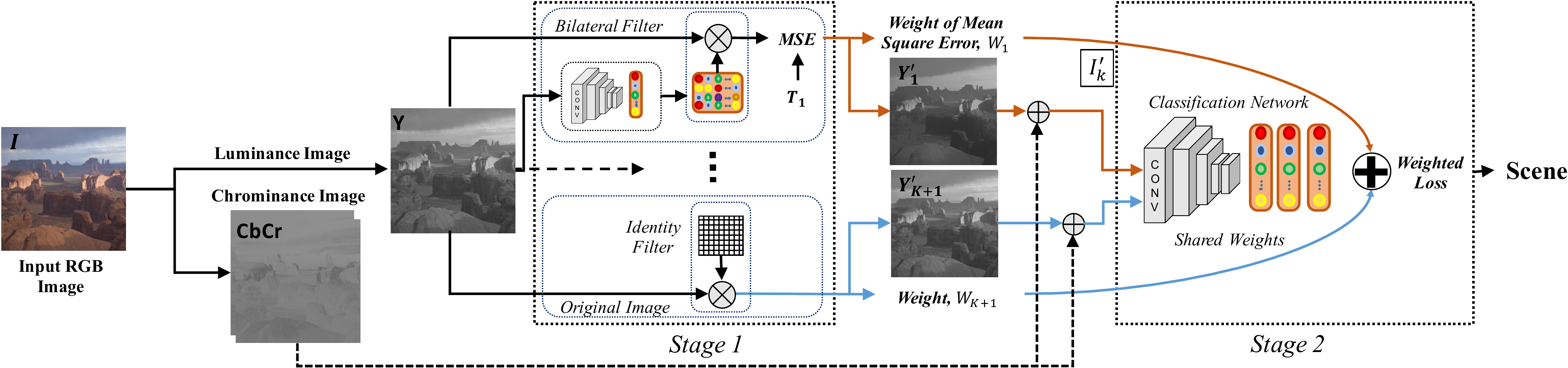}
\caption{\small{\textbf{(Dyn-CNN)} In this architecture, similar to Sec.~\ref{subsec:app1} (Figure~\ref{fig:2}), we show end-to-end joint \textbf{learning of multiple filters}. The individual softmax scores are combined in a weighted way in the CNN.  There is \textbf{adaptation} of weights of the filters.}}\vspace{-0.5cm}
\label{fig:4}
\end{figure*}
\subsection{Static Filters for Classification} \label{subsec:app2}

Here, we show how to integrate the pre-learned enhancement filters obtained from the first approach. 
For each image in the train set, we obtain a dynamic filter using our first approach. The static filter is computed by taking a mean of all these dynamic filters. The extracted static filters are convolved with the input luminance $Y$ component of the RGB image $I$, and the chrominance component is added  and then the image is transformed back to RGB $I^{'}$, which is then fed into the classification network. Figure~\ref{fig:3} shows the schematic layout of the whole architecture.

\textbf{First stage (enhancement stage):} We begin by extracting the pre-trained filters for $K$ image enhancement methods learned from the first approach. Given an input luminance image $Y$, these $f_{\Theta,k}$ filters are convolved with the input image to generate $Y^{'}_{k}$ enhanced images as $Y^{'}_{k}=f_{\Theta,k}(Y), k \in K$. We also include an identity filter ($K$+1) to generate the original image, as some learned enhancements may perform worse than the original image itself. We then investigate two different strategies to weight $W_{k}$ the enhancement methods: (1) giving equal weights with value equal to 1/$K$, and (2) giving weights on the basis of MSE, as discussed in Sec.~\ref{subsec:app3}. 

The output of this stage is a set of enhanced luminance images and their corresponding weights indicating the potential importance for pushing to the next stage of the classification pipeline. Chrominance is then recombined, and the images are transformed back to RGB, $I^{'}_{k}$.

\textbf{Second stage (classification stage):}
The enhanced images $I^{'}_{k}$ for $K$ image enhancement methods and original image are fed as an input to the classification network one by one sequentially, with class labels and their weights $W_{k}$ indicating the importance of the enhancement method for the input image. Similar to the last approach, the network parameters of the fully-connected layer and $C$-way classification layer are fine-tuned using a pre-trained network in an end-to-end learning approach.

\textbf{End-to-end training:}
The loss of the network training is the weighted  $W_{k}$ sum of the individual softmax losses $\Lb_{k}$ term. The weighted loss is given as: 
\begin{equation}
Loss_{Stat} = \sum_{k=1}^{K+1}W_{k} \Lb_{k}(\mathbf{P},\mathbf{y})
\end{equation}
where $W$ is the weight indicating the importance of the $K$  enhancement method, where $W_{K+1}=1$ for original RGB image, contributing to the total loss of the whole pipeline.

\subsection{Multiple Dynamic Filters for Classification} \label{subsec:app3}

Here, we recycle the architectures from  Sections~\ref{subsec:app1}-\ref{subsec:app2}. Figure~\ref{fig:4} shows the schematic layout of the whole architecture. Our architecture uses the similar architecture proposed in Sec.~\ref{subsec:app1}; we dynamically generate  $K$ filters using $K$ enhancement networks, one for each enhancement method. In this proposed architecture, the loss associated with Stage 1 is the MSE between the predicted output images $Y^{'}_{k}$ and the target output images $T_{k}$.

For computing the weights of each enhancement method, the MSE for the enhanced images are transformed to weights $W_{k}$ by comparing their relative strengths as: $W_{k}=MSE_{k}/\sum_{m=1}^{K}MSE_{m}$, followed by $W_{k}=(W_{k}-max(W))/(min(W)-max(W))$. Since now the $min(W)$ is zero, in order to avoid giving zero-weight to one of the enhancement methods, we subtract the \textit{second-$min(W)/2$} from $W$ and add the \textit{second-$min(W)$} to the $W_{k}$ with $min(W)$. Finally, we scale the weights $W_{k}=W_{k}/\sum_{m=1}^{K}W_{m}$ with the constraint that the sum of the weights for all $K$ methods should be equal to $1$. The enhanced images with the smallest errors obtain the highest weight, and vice versa. In addition, we also compare against giving equal weights to all enhancement methods. Of both weighting strategies, MSE-based weighting yielded the best results, and was therefore selected as the default. Note that we also include the original image by simply convolving it with an identity filter ($K$+1) similar to approach 2: the weight for the RGB image is set to $1$, i.e.  $W_{K+1}=1$. During training, the weights are estimated by cross-validation on the train/validation set, while for the testing phase, we use these pre-computed weights. Further, we observed that training the network without regularization of weights has prevented the model from converging throughout the learning, and led to overfitting with significant drop in performance.

\textbf{End-to-end training:}
Finally, we now extend the loss of approach 2, by adding an MSE term for joint optimization of $K$ enhancement networks with a classification objective. We learn all parameters of the network jointly in an end-to-end fashion. The weighted loss is \textit{sample-specific}, and is given as:
\begin{equation}
\begin{split}
Loss_{Dyn} = \sum_{k=1}^{K}MSE_{k}(T_{k},Y^{'}_{k}) + \sum_{k=1}^{K+1}W_{k}\Lb_{k}(\mathbf{P},\mathbf{y})
\end{split}
\end{equation}

We believe training our network in this manner, offers a natural way to encourage the filters to apply a transformation that enhances the image structures for an accurate classification, as the classification network is regularized via enhancement networks. Moreover, joint optimization helps minimize the overall cost function of the whole architecture,  hence leading to better results.

\section{Experiments} \label{sec:exp}

In this section, we demonstrate the use of our enhancement filtering technique on four very different image classification tasks. First, we introduce the dataset, target output data generation,  and implementation details, and explore the design choices of the proposed methods. Finally, we test and compare our proposed method with baseline methods and other current ConvNet architectures. Note that, the purpose of this paper is to improve the baseline performance of generic CNN architectures using an \textit{add-on} enhancement filters, and not to compete against the state-of-the-art methods.

\subsection{Datasets}  

We evaluate our proposed method on four visual recognition tasks: fine-grained classification using CUB-200-2011 CUB)~\cite{cubdataset}, object classification using PASCAL-VOC2007 (PascalVOC)~\cite{pascalvocdataset},  scene recognition using MIT-Indoor-Scene (MIT)~\cite{mitdataset}, and texture classification using Describable Textures Dataset (DTD)~\cite{dtddataset}. Table~\ref{table:databases} shows the details of the datasets. For all of these datasets, we use the standard training/validation/testing protocols provided as the original evaluation scheme and report the classification accuracy.

\begin{table}[b] 
\begin{center}
\resizebox{6.5cm}{!} {
\begin{tabular}{ l |   c   c   c     }
\hline
Data-set &  \# train img & \# test img & \# classes\\
\hline
\hline
CUB~\cite{cubdataset}  & 5994 &5794& 200 \\
PascalVOC~\cite{pascalvocdataset}  & 5011&4952& 20\\

MIT~\cite{mitdataset}   &4017&1339&67 \\

DTD~\cite{dtddataset}  &1880&3760&47 \\
\hline
\end{tabular}}
\end{center}\vspace{-0.3cm}
\caption{\small{Details of the training and test set for datasets.}}
\label{table:databases}
\end{table}

\subsection{Target Output Data} \label{subsec:gt}
We generate target output $T$ images for five (i.e., $K$=5) enhancement methods $E$: (1) weighted least squares~(WLS) filter~\cite{wls}, (2) bilateral filter (BF)~\cite{fastbilateral,bilateral}, (3) image sharpening filter (Imsharp), (4) guided filter (GF)~\cite{guided}, and (5) histogram equalization (HistEq). Given an input image, we first transform the RGB color space into a luminance-chrominance color space, and then apply these enhancement methods on the luminance image to obtain an enhanced luminance image. This enhanced luminance image is then used as the target image for training.  We used default parameters for WLS and Imsharp, and for BF, GF and HistEq parameters are adapted to each image, thus requiring no parameter setting. For comprehensive discussion, we refer the readers to~\cite{fastbilateral,wls,guided}. The source code for fast BF~\cite{fastbilateral}, WLS~\cite{wls} is publicly available, and others are available in the Matlab framework.

\subsection{Implementation Details}
We use the MatCovNet and Torch frameworks, and all the ConvNets are trained on a TitanX GPU. Here we discuss the implementation details for ConvNet training  (1) with dynamic enhancement filter networks, (2) with static enhancement filters, and (3) without enhancement filters as a classic ConvNet training scenario.

We evaluate our design on AlexNet~\cite{alexnet}, GoogleNet~\cite{googlenet}, VGG-VD~\cite{vgg}, VGG-16~\cite{vgg}, and BN-Inception~\cite{bninception}. In each case, the  models are pre-trained on the ImageNet~\cite{imagenet} and then fine-tuned on the target datasets. To fine-tune the network, we replace the 1000-way classification layer with a $C$-way softmax layer, where $C$ is the number of classes in the target dataset.  For fine-tuning the different architectures depending on the dataset about 60-90 epochs (batch size 32) were used, with a scheduled learning rate decrease, starting with a small learning rate $0.01$. All ConvNet architectures are trained with identical optimization schemes, using SGD optimizer with a fixed weight decay of $5\times10^{-4}$ and a scheduled learning rate decrease. We follow two steps to fine-tune the whole network. First, we fine-tune (last two fc layers) the ConvNet architecture using RGB images, and then embed it in Stat/Dyn-CNN for fine-tuning the whole network with enhancement filters, by setting a  small learning rate for all layers except the last two fc layers, which have a high learning rate. Specifically, for example, in BN-Inception the network requires a fixed input size of $224 \times 224$. The images are mean-subtracted before network training. We apply data augmentation~\cite{alexnet,vgg} by cropping the four corners, center, and their x-axis flips, along with color jittering (and the cropping procedure repeated for each of these) for network training. Ahead we provide more details for ConvNet training using BN-Inception.

$-$\textbf{Dynamic enhancement filters (Dyn-CNN):}
The enhancement network consists of  $\sim$570k learnable model parameters, with the last fully-connected layer (i.e., dynamic filter parameters) containing 36 neurons - that is, filter-size $6\times6$.  We initialize the enhancement networks' model parameters randomly, except for the last fully-connected layer, which is initialized to regress the identity transform (zero weights, and identity transform bias), suggested in~\cite{stn}. We initialize the learning rate with $0.01$ and decrease it by a factor of 10 after every 15k iterations. The maximum number of iterations is set to 90k. In terms of computation speed, the training enhancement network along with BN-Inception  takes approx. 7\% more training time for network convergence in comparison to BN-Inception for approach 1 (Sec.~\ref{subsec:app1}). We use five enhancement networks for generating five enhancement filters (one for each method) for approach 3 (Sec.~\ref{subsec:app3}). We also include original RGB image too.

$-$\textbf{Without enhancement filters (FC-CNN):}
Similar to classical ConvNets' fine-tuning scenario, we replace the last classification layer of a pre-trained model with a $C$-way classification layer before fine-tuning. The fully connected layers and the classification layer are fine-tuned.  We initialize the learning rate with $0.01$ and decrease it by a factor of 10 after every 15k iterations. The maximum number of iterations is set to 45k. 

$-$\textbf{Static enhancement filters (Stat-CNN):}
Similar to FC-CNN, here, we have five enhanced images for five static filters and original RGB image as the sixth input that are fed as  input to the ConvNets for network training. In practice, the static filters for image enhancement are very low-complex operations. The optimization scheme used here is the same as FC-CNN. We use all the five static learned filters for approach 2 (Sec.~\ref{subsec:app2}).

\noindent
\textbf{Testing:}  
As previously mentioned, the input RGB image is transformed into luminance-chrominance color space, and then the luminance image is convolved with the enhancement filter, leading to an enhanced luminance image. Chrominance is  then recombined to the enhanced luminance image and the image is transformed back to RGB.  For ConvNet testing,  an input frame with either be an RGB image or an enhanced RGB image using the static or dynamic filters is fed into the network. In total, five enhanced images (one for each filter) and the original RGB image are fed into the network, sequentially. For final image label prediction, the predictions of all images are combined through a weighted sum, where the pre-computed weights $W$ are obtained from Dyn-CNN.

\subsection{Fine-Grained Classification} \label{subsec:cubeval}
In this section, we use CUB-200-2011~\cite{cubdataset}  dataset as a test bed to explore the design choices of our proposed method, and then finally compare our method with baseline methods and the current methods. 

\noindent
\textbf{Dataset:} CUB~\cite{cubdataset} is a fine-grained bird species classification dataset. The dataset contains 20 bird species with 11,788 images. For this dataset, we  measure the accuracy of predicting a class for an image. 

\noindent
\textbf{Ablation study:} Here we explore four aspects of our proposed method: (1) the impact of different filter size;  (2) the impact of each enhancement method, separately; (3) the impact of weighting strategies;  and (4) the impact of different ConvNet architectures.  

$-$\textbf{Filter size:} In our experiment, we explore three different filter sizes. Specifically, we implement the enhancement network as a few convolutional and  fully-connected layers with the last-layer containing (1) 25 neurons ($f_{\Theta}$ is an enhancement filter of size $5 \times 5$),  (2) 36 neurons ($6 \times 6$), and (3) 49 neurons ($7\times7$). From the literature~\cite{dfn,stn}, we exploited the insights about good filter size.  The filter size  determines the receptive field and is application dependent. We found that a filter size $>$ $7\times7$ produces smoother images, and thus drops the classification performance by approx. 2\% (WLS: $68.73 \rightarrow 66.84 $) in comparison to a filter size of $6\times6$. Similar was the case with a filter size $<$ $5\times5$, where correct enhancement was not transferred, leading to a drop in performance  by approx. 3\% (WLS: $68.73 \rightarrow 65.9$). We found that the filter size $6\times6$ learned the expected transformation, and applied the correct enhancement to the input image with sharper preserved edges.

$-$\textbf{Enhancement method ($E$):} Here, we compare the performance of individual enhancement methods in three aspects: (1) We employ AlexNet~\cite{alexnet} pre-trained on ImageNet~\cite{imagenet} and fine-tuned (last two fc layers) on CUB for each ground-truth enhancement method separately (GT-EMs). (2) Using the pre-trained RGB AlexNet model on CUB from (1), we fine-tune the whole model with GT-EMs, by setting a small learning rate for all layers except the last two fc layers, which have a high learning rate. This slightly improves the performance of the pre-trained RGB model by a small margin. (3) Similar to (2), but here we fine-tune the whole model using approach 1 (Sec.~\ref{subsec:app1}). We can see that our dynamic enhancement approach improves the performance by a margin of  $\sim$1-1.5\% in comparison to a generic network when fine-tuned on RGB images only. In Table~\ref{table:enhanceeval}, we summarize the results. 

\begin{figure}[b]
 \centering
 \includegraphics[width=0.99\columnwidth]{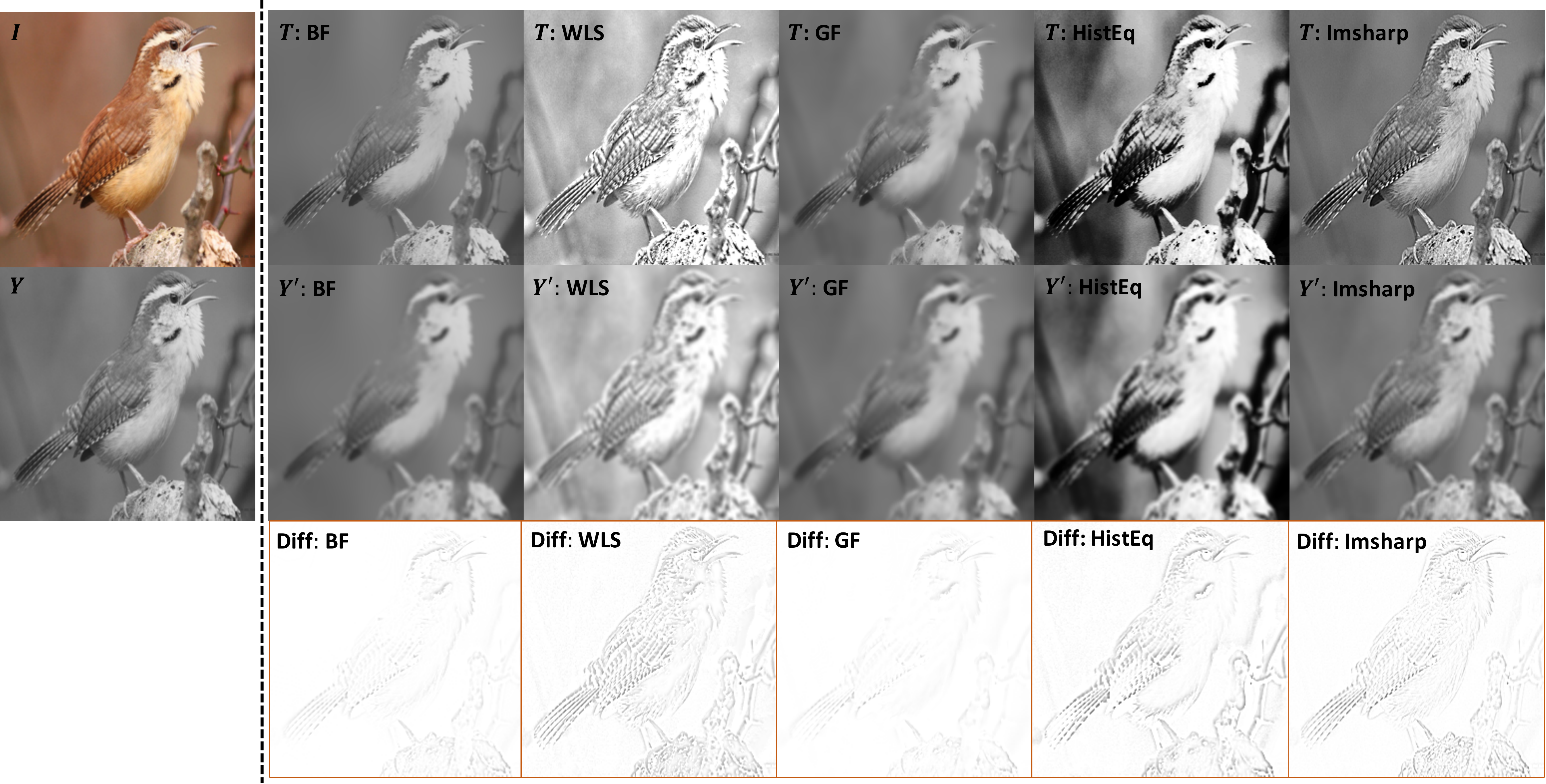}
 \caption{\small{\textbf{Qualitative results: CUB}. Comparison between the target image $T$, enhanced luminance image $Y^{'}$, and compliment of difference image (diff=$T$-$Y^{'}$) obtained using approach 1 (Sec.~\ref{subsec:app1}) for all enhancement methods.}} \vspace{-0.3cm}
  \label{fig:5}
 \end{figure}

In Fig.~\ref{fig:5}, as an example we show some qualitative results for the difference in textures that our enhancement method extracts from the GT-EMs, which is primarily responsible for improving the classification performance.

\begin{table}[t] 
\begin{center}
\resizebox{8.5cm}{!} {
\begin{tabular}{ l |   c |  c   c c c c | c     }
\hline
   & RGB & BF & WLS & GF & HistEq & Imsharp & LF\\
\hline
(a) GT-EMs 		& 67.3~\cite{Bubblenet}  	&  66.93 & 67.34 & 67.12  & 66.41 & 66.74 	&  70.14	\\
(b) RGB: GT-EMs  		& $-$   	&  67.16 & 67.41 & 67.37  & 66.58 & 66.87 	&   71.28	\\
(c) \textbf{Ours} (Sec.~\ref{subsec:app1})  			& $-$ 	&  \textbf{68.21}  & \textbf{68.73} & \textbf{68.5}  & \textbf{67.62} & \textbf{67.86}  & \textbf{72.16}	\\
\hline
\end{tabular}}
\end{center}\vspace{-0.3cm}
\caption{\small{Individual accuracy (\%) performance comparison of all the enhancement methods $E$ using AlexNet on CUB, where LF is \textit{late-fusion} as averaging of scores for the 5 enhancement methods.}}\vspace{-0.5cm}
\label{table:enhanceeval}
\end{table}

\begin{table}[b] 
\begin{center}
\resizebox{8cm}{!} {
\begin{tabular}{ c |      c   c c c c | c    }
\hline
  &  BF & WLS & GF & HistEq & Imsharp & RGB\\
\hline
$W$  & 0.23$\pm$0.05  & 0.25$\pm$0.04  & 0.24$\pm$0.03 & 0.13$\pm$0.03  & 0.17$\pm$0.05 	& 1.0\\
\hline
\end{tabular}}
\end{center}\vspace{-0.3cm}
\caption{\small{Relative comparison of the strength of weights $W$ for each enhancement method estimated by cross-validation on the training set of CUB using Dyn-CNN with BN-Inception, where $W$ for RGB image by default is set to 1.}}
\label{table:weighteval}
\end{table}

$-$\textbf{Weighting strategies:}  Combining the enhancement methods in a late-fusion (LF) as an averaging of the scores gives further improvements, shown in Table~\ref{table:enhanceeval}. With this observation, we realized a more effective weighting strategy should be applied, such that more importance could be given to better methods for combining.  In our evaluation, we explore two weighting strategies (1) giving equal weights $W_{k}$ with value equal to $1/K$ - that is, 0.2 for $K$=5, and (2) weight computed on the basis of $MSE$, estimated by cross-validation on the training set, shown in Table~\ref{table:weighteval}.  

Table~\ref{table:weighteval1} clearly shows that weighting adds a positive regularization effect. We found that training the network with regularization of MSE loss prevents the classification objective from divergence throughout the learning. Table~\ref{table:weighteval} shows that in Dyn-CNN the weight of each enhancement filter relates very well to that of its individual performance shown in Table~\ref{table:enhanceeval}. We observe that the MSE-based weighting performs the best. Therefore, we choose that as a default weighting method. 

\begin{table}[t] 
\begin{center}
\resizebox{6cm}{!} {
\begin{tabular}{ l |   c   c   c c c c     }
\hline
 					&   Stat-CNN (ours) &  Dyn-CNN (ours)  \\
\hline
$W$: Averaging 				& 83.19	&  {85.58}\\
$W$: MSE-based 				& 83.74	&  \textbf{86.12}\\
\hline
\end{tabular}}
\end{center}\vspace{-0.3cm}
\caption{\small{Accuracy (\%) performance comparison of the weighting strategies using BN-Inception on CUB.}}\vspace{-0.5cm}
\label{table:weighteval1}
\end{table}

$-$\textbf{ConvNet architectures:} Here, we compare the different ConvNet architectures. Specifically, we compare AlexNet~\cite{alexnet}, GoogLeNet~\cite{googlenet}, and BN-Inception~\cite{bninception}. Among all architectures shown in Table~\ref{table:convneteval}, BN-Inception exhibits the best performance in terms of classification accuracy in comparison to others. Therefore, we choose BN-Inception as a default architecture for this experiment. 

\begin{table}[b] 
\begin{center}
\resizebox{8cm}{!} {
\begin{tabular}{ l |   c   c   c c c c     }
\hline
 					&   FC-CNN &  Stat-CNN (ours)& Dyn-CNN (ours) \\
\hline
AlexNet 		 		& 67.3~\cite{Bubblenet} 	&  68.52  	& 73.57\\
GoogLeNet 			& 81.0~\cite{Bubblenet}	&  82.35	&  84.91\\
BN-Inception 			& 82.3~\cite{stn}		&  83.74	&  \textbf{86.12}\\
\hline
\end{tabular}}
\end{center}\vspace{-0.3cm}
\caption{\small{Accuracy (\%) performance comparison of different architectures on CUB.}}\vspace{-0.5cm}
\label{table:convneteval} 
\end{table}

\noindent
\textbf{Results:}
In Table~\ref{table:cubeval},  we  explore our static and dynamic CNNs with the current methods. We consider BN-Inception using our two-step fine-tuning scheme with Stat-CNN and Dyn-CNN. We can notice that Dyn-CNN improves the generic BN-Inception performance by \textbf{3.82\%} ($82.3 \rightarrow {86.12}$) using  image enhancement.  Our EnhanceNet takes a constant time of only \textbf{8 ms} (GPU) to generate all enhanced images altogether, in comparison to generating ground-truth target images, which is very time-consuming and takes $\sim$1-6 seconds for each image/method: BF, WLS, and GF. Testing time for the whole model is: EnhanceNet (8 ms) plus ClassNet (inference time for the architecture used).

Further, we extend the baseline $2 \times$ST-CNN~\cite{stn} to include static (Sec~\ref{subsec:app2}) and dynamic filters (Sec.~\ref{subsec:app3}) immediately following the input, with the weighted loss. In reference to ST-CNN work~\cite{stn}, we evaluate the methods, keeping the training and evaluation setup the same for a fair comparison. Our results indicate that Dyn-CNN improves the performance by  \textbf{3.81\%} ($83.1 \rightarrow {86.91}$). Furthermore, our Stat-CNN with static filters is competitive too, and performs 1.15\% better than $2\times$ST-CNN~\cite{stn}. This means that the static filters when dropped into a network can perform explicit  enhancement of features, and thus gains in accuracy are expected in any ConvNet architecture.

\begin{table}[t] 
\begin{center}
\resizebox{8cm}{!} {
\begin{tabular}{ l |   c   c   c c c c     }
\hline
 					&   FC-CNN &  Stat-CNN (ours) & Dyn-CNN (ours) \\
\hline
$4\times$ ST-CNN~\cite{stn}: 448px 			& 84.1~\cite{stn}		&  $-$	& $-$\\
\hline
BN-Inception 						& 82.3~\cite{stn}		&  83.74	&  86.12\\
$2\times$ ST-CNN~\cite{stn} 			&83.1~\cite{stn}		&{84.25}	&  \textbf{86.91}\\
\hline
\end{tabular}}
\end{center}\vspace{-0.3cm}
\caption{\small{\textbf{Fine-grained classification (CUB)}. Accuracy (\%) performance comparison of Stat-CNN and Dyn-CNN with baseline methods and previous works on CUB.}}\vspace{-0.2cm}
\label{table:cubeval}
\end{table}

\subsection{Object Classification} \label{subsec:pascalvoceval}

\noindent
\textbf{Dataset:} The PASCAL-VOC2007~\cite{pascalvocdataset} dataset contains 20 object classes with 9,963 images that contain a total of 24,640 annotated objects.  For this dataset, we  report the mean average precision (mAP), averaged over all classes.

\noindent
\textbf{Results:} In Table~\ref{table:pascalvoceval}, we show the results. Dyn-CNN is \textbf{4.58/6.16}\% better than Stat-CNN/FC-CNN using AlexNet, and  \textbf{2.43/3.5}\% using VGG-16. One can observe that for a smaller network, AlexNet shows more improvement in performance in comparison to the deeper: VGG-16 network. Also, Stat-CNN is 1.58/1.07\% better than FC-CNN using AlexNet/VGG-16.  Furthermore, Bilen et al.~\cite{bilen2016weakly} with 89.7\% mAP performs \textbf{3.1\%}  ($89.7 \leftarrow {92.8}$) lower than Dyn-CNN using VGG-16.

\begin{table}[t] 
\begin{center}
\resizebox{8cm}{!} {
\begin{tabular}{ l   | l |   c  c c   }
\hline
 Dataset  &   ConvNet &FC-CNN &  Stat-CNN & Dyn-CNN \\
					&		&    		 &  (ours)		 & (ours) 		 \\
\hline
\hline
PasclVOC (mAP) &AlexNet 		&76.9~\cite{tang2016deep}  		&78.48	& 83.06 \\
PasclVOC (mAP) &VGG-16 		&89.3~\cite{vgg}				&{90.37} 	&\textbf{92.8}\\
\hline
MIT (Acc.) &AlexNet 				&56.79~\cite{zhou2016places}		&58.24	& 62.9 \\
MIT (Acc.) & VGG-16 			&64.87~\cite{zhou2016places}		&{65.94}	& \textbf{68.67} \\
\hline
DTD (mAP)  & AlexNet 			& 61.3~\cite{cimpoi2016deep}		&62.9 	& 67.81\\
DTD (mAP) & VGG-VD 			&67.0~\cite{cimpoi2016deep}		&{69.12}	& \textbf{71.34}\\
\hline
\end{tabular}}
\end{center}\vspace{-0.3cm}
\caption{\small{\textbf{Performance comparison in \%}. The table compares FC-CNN, Stat-CNN, and Dyn-CNN on AlexNet and VGG networks trained on ImageNet and fine-tuned on target datasets using the standard training and testing sets. \iffalse The table shows detailed comparison for Pascal-VOC2007~\cite{pascalvocdataset}, MIT-Indoor~\cite{mitdataset}, and DTD~\cite{dtddataset}. \fi}}\vspace{-0.5cm}
\label{table:pascalvoceval}
\end{table}

\subsection{Indoor Scene Recognition} \label{subsec:miteval}

\noindent
\textbf{Dataset:} The MIT-Indoor scene dataset (MIT)~\cite{mitdataset} contains a total of  67 indoor scene categories with 5,356 images. For this dataset, we  measure the accuracy of predicting a class for an image.  

\noindent
\textbf{Results:} In Table~\ref{table:pascalvoceval}, we show the results. As expected and previously observed, Dyn-CNN is \textbf{4.66/6.11}\% better than Stat-CNN/FC-CNN using AlexNet, and  \textbf{2.73/3.8}\% using VGG-16. 

\subsection{Texture Classification} \label{subsec:dtdeval}

\noindent
\textbf{Dataset:} The Describable Texture Datasets (DTD)~\cite{dtddataset} contains 47 describable attributes with 5,640 images.  For this dataset, we  report the mAP, averaged over all classes.

\noindent
\textbf{Results:} In Table~\ref{table:pascalvoceval}, we show the results. The story is similar  to our previous observation: Dyn-CNN outperforms Stat-CNN and FC-CNN by a significant margin. Surprisingly, it is interesting to see that Dyn-CNN shows a significant improvement of \textbf{6.51/4.34\%} in comparison to FC-CNN using AlexNet/VGG-VD.

\section{Concluding Remarks} \label{sec:conc}
In this paper, we propose a unified CNN architecture that can emulate a range of enhancement filters with the overall goal to improve image classification in an end-to-end learning approach. We demonstrate our framework on four benchmark datasets: PASCAL-VOC2007, CUB-200-2011,  MIT-Indoor Scene, and Describable Textures Dataset. In addition to improving the baseline performance of  vanilla CNN architectures on all datasets, our method shows promising results in comparison to the state-of-the-art using our static/dynamic enhancement filters. Also, our enhancement filters can be used with any existing networks to perform explicit enhancement of image texture and structure features, giving CNNs higher-quality features to learn from, which in turn can lead to more accurate classification.

We believe our work opens many possibilities for further exploration. In future work, we plan to further investigate more enhancement methods as well as more complex loss functions which are appropriate for image enhancement tasks.

\noindent
\small{\textbf{Acknowledgements:} This work is supported by the DFG, a German Research Foundation - funded PLUMCOT project, the CR-REF VISTA  programme, and an NSERC Discovery Grant.}

{\small
\bibliographystyle{ieee}
\bibliography{egbib}
}

\end{document}